\documentclass[conference]{IEEEtran}
\usepackage{graphicx}
\usepackage{epstopdf}
\usepackage{hyperref}
\def\R2Lurl#1#2{\mbox{\href{#1}{\tt #2}}}
\begin{document}
\title{Edit wars in Wikipedia}

\author{\IEEEauthorblockN{R\'obert Sumi, Taha Yasseri, Andr\'as Rung}
\IEEEauthorblockA{Institute of Physics\\
Budapest University of Technology and Economics\\
H-1111 Budapest Budafoki u 8\\
Email: \{rsumi,yasseri,runga\}@phy.bme.hu}
\and
\IEEEauthorblockN{Andr\'as Kornai\hfill J\'anos Kert\'esz}
\IEEEauthorblockA{Institute of Mathematics and Institute of Physics}
Budapest University of Technology and Economics\\
H-1111 Budapest Budafoki u 8\\
Email: kornai@math.bme.hu, kertesz@phy.bme.hu}


%

\maketitle

\begin{abstract} We present a new, efficient method for automatically
detecting severe conflicts, `edit wars' in Wikipedia and evaluate this method
on six different language Wikipedias. We discuss how the number of edits and reverts
 deviate in such pages from those following the general workflow, and argue that earlier work
has significantly over-estimated the contentiousness of the Wikipedia editing
process.
\end{abstract}

\section{Introduction}

The development of Wikipedia (WP) articles is not always a peaceful and
collaborative process. This has long been recognized by the WP community, which
calls extreme cases of disagreement over the contents of an article an
\R2Lurl{http://en.wikipedia.org/wiki/Edit_war}{edit war},\footnote{In the
  electronic version of this paper hyperlinks are direct references to WP.  In
  the printed version, these are given in {\tt typewriter font}}. WP has developed
specific guidelines for dealing with edit warring, such as the
\R2Lurl{http://en.wikipedia.org/wiki/Wikipedia:Edit_warring\#The_three-revert_rule}{three
  revert rule}, offers a variety of tags to warn about
\R2Lurl{http://en.wikipedia.org/wiki/Wikipedia:Template_messages\#Disputes_and_warnings}{disputes}, and even has a humorous listing of the
\R2Lurl{http://en.wikipedia.org/wiki/Wikipedia:Lamest_edit_wars/Ethnic_feuds}{lamest edit wars}.

Perhaps the easiest way human readers can detect pages
affected by warring (in the English WP, which is the one discussed unless
explicitly stated otherwise)  is to read through the {\it discussion page} (also known
as the {\it talk page}) associated to each content page looking for telltale
signs such as notices requesting cleanup, swearwords and name-calling. When
the discussion grows heated, the length of the talk page can exceed the length
of the article many times over, so that older discussions must be
archived. Another way to detect controversy is to view the {\it history} of
the page, which can show many war-like acts, in particular editors {\it
  reverting} the work of other editors.

Schneider et al \cite{schneider:2010} estimate that among highly edited or highly viewed
articles (these notions are strongly correlated, see \cite{Ratkiewicz2010}) about 12\% of discussions is devoted to reverts and
vandalism, suggesting that the WP development process is highly
contentious. In fact, once the great bulk of WP articles is considered, we
find the editorial process far more peaceful: as we shall see, around 24k
articles, i.e. less than 1\% of the 3m articles available in the November 2009
English WP dump, can be called controversial. To sustain such far-reaching conclusions we can no longer
rely on manual checking, so our primary interest is with the automatic
detection of edit wars. Since our interest is with the entire WP process, of
which the English WP is just the largest (and most mature) instance, we are
primarily interested in language- and culture-independent methods that can be
applied uniformly across the range of WPs. Therefore, our methods are
based entirely on the history page, as opposed to the more human-readable talk
page.

Previous works (including our own) aimed at the automatic detection of
editorial conflict and edit wars is summarized in Section~\ref{sec:detect}. In Section~\ref{sec:other}
we discuss other indicators of controversiality and evaluate these in comparison to ours.
We offer our conclusions in Section~\ref{sec:conc}.

\section{Automatic conflict detection}\label{sec:detect}

Conflicts in WP were studied already both on the article and on the user
level. Kittur~\cite{Kittur2007} et al. computed article controversy from
different page metrics (number of reverts, number of revisions etc.), Vuong et
al.~\cite{Voung2008} counted the number of deleted words between users and
used their ``Mutual Reinforcement Principle" to measure how controversial a
given article is. Both teams counted how many times dispute tags
appeared in the history of an article, and used this as ground truth. While
this is an excellent test in one direction (certainly recognition of
controversiality by the participants is as good as the same recognition coming
from an outsider), it is too narrow, as there can be quite significant wars
that the community is unaware of or at least do not tag, as, e.g., in the
articles on {\tt Gda\'{n}sk} or {\tt Euthanasia}. Note that by applying more
lax criteria (i.e. not requiring the presence of overt conflict tags) our
method will, if anything, overestimate the extent of controversy,
strengthening our conclusion that there is much {\it
  less} conflict in WP than appears from sampling highly edited/viewed pages.

There are several papers which try to measure the negative links between WP
editors in a given article and, based on this, attempt to classify editors
into groups. The main idea of the method used by Kittur et
al.~\cite{Kittur2007,Suh2007} is to count how many times an editor pair
reverted each other. The more two editors reverted each other, the larger the
conflict between them. As we shall see shortly, reverts are indeed central to
the assessment of controversiality, but one needs to take into account not
just the number of (presumably hostile) interactions, but also the seniority
of the participants.  Brandes et al.~\cite{Brandes2008} assumed that users who
do not agree with each other react very fast to edits by the others. The
reciprocal value of the time elapsed between two consecutive edits increases
the controversy between the two authors.  In a more recent paper
Brandes~\cite{Brandes2009} counted the number of deleted words between editors
and used this as a measure of controversy.  West et. al. and also Adler
et. al. have developed vandalism detection methods based on temporal patterns
of edits \cite{west2010,adler2010}. In both works the main assumption is that
offensive edits are reverted much faster than normal edits, therefore by
considering the time interval between an arbitrary edit and its subsequent
reverts, one can classify vandalized versions with high precision.

Our own work (for a preliminary report, see \cite{sumi2011}) was seeded by a
manual sample of 40 articles, 20 selected for high controversiality, and 20
for low. Table~\ref{table:revertdet} summarizes the number of reverts as detected in the text and
in the comments\footnote{Each edit could be accompanied by a descriptive comment as {\it Edit summary}.} (most reverts are detected by both methods).

\begin{center}
\begin{table}[h!]
\caption{Number of reverts detected. The upper part corresponds to a group of pages with severe conflicts (except those
in {\it italics}); below the horizontal line there are peaceful pages (except those in {\it italics}).}
  \begin{tabular}{ | l | l | p{1.2cm} | p{4cm} |}
    \hline
Both txt & Only & Only & Article \\
and cmt & in txt & in cmt & titlee \\ \hline
4103 & 930 & 328 & Global warming \\
2375 & 478 & 142 & Homosexuality \\
1847 & 617 & 201 & Abortion \\
1494 & 260 & 35 & {\it Benjamin Franklin} \\
1425 & 437 & 130 & Elvis Presley \\
1396 & 233 & 67 & Nuclear power \\
1298 & 536 & 104 & Nicolaus Copernicus \\
1071 & 211 & 51 & Tiger \\
1036 & 248 & 58 & Euthanasia \\
937 & 204 & 58 & Alzheimer's disease \\
870 & 192 & 50 & Gun politics \\
836 & 172 & 23 & {\it Sherlock Holmes} \\
689 & 213 & 49 & Arab-Israeli conflict \\
659 & 496 & 138 & Israel and the apartheid analogy \\
652 & 387 & 88 & Liancourt Rocks \\
642 & 250 & 39 & Schizophrenia \\
516 & 164 & 472 & Gaza war \\
431 & 186 & 30 & 1948 Arab-Israeli war \\
416 & 73 & 9 & {\it Pumpkin} \\
380 & 284 & 58 & Gda\'{n}sk \\
318 & 158 & 20 & SQL \\
    \hline
162 & 24 & 10 & Password \\
116 & 26 & 3 & Henry Cavendish \\
109 & 29 & 4 & Pension \\
81 & 29 & 4 & Mexican drug war \\
74 & 37 & 10 & {\it Hungarians in Romania} \\
70 & 14 & 4 & Markov chain \\
70 & 12 & 1 & Mentha \\
47 & 20 & 6 & Foucault pendulum \\
40 & 5 & 6 & Indian cobra \\
32 & 15 & 1 & Harmonium \\
30 & 9 & 1 & Infrared photography \\
29 & 4 & 1 & Bohrium \\
24 & 34 & 5 & {\it Anyos Jedlik}  \\
11 & 6 & 2 & Hungarian forint \\
10 & 3 & 1 & Hendrik Lorentz \\
9 & 3 & 1 & 1980s oil glut \\
7 & 1 & 0 & Deutsches Museum \\
4 & 0 & 0 & Ara (genus) \\
0 & 0 & 0 & Schlenk flask \\
    \hline
    \end{tabular}
\label{table:revertdet}
\end{table}
\end{center}

Given the number and distribution of false positives
and negatives (typeset in italics) it is clear from Table~\ref{table:revertdet} that the raw
revert statistics do not yield a clear cutoff-point we could use to
distinguish controversial from non-controversial articles. Rather than
building a complex but arbitrary formula that includes different factors that are expected to
correlate with controversiality, our goal is to base the decision on very few
parameters -- ideally, just one.

Let $\dots, i-1, i, i+1, \dots, j-1, j, j+1, \dots$ be stages in the history
of an article. If the text of revision $j$ coincides with the text of revision
$i-1$, we considered this a revert between the editor of revision $j$ and $i$
respectively.  We are interested in disputes where editors have different
opinions about the topic, and do not reach consensus easily. 

\begin{figure}[h!]
\includegraphics[width=52mm]{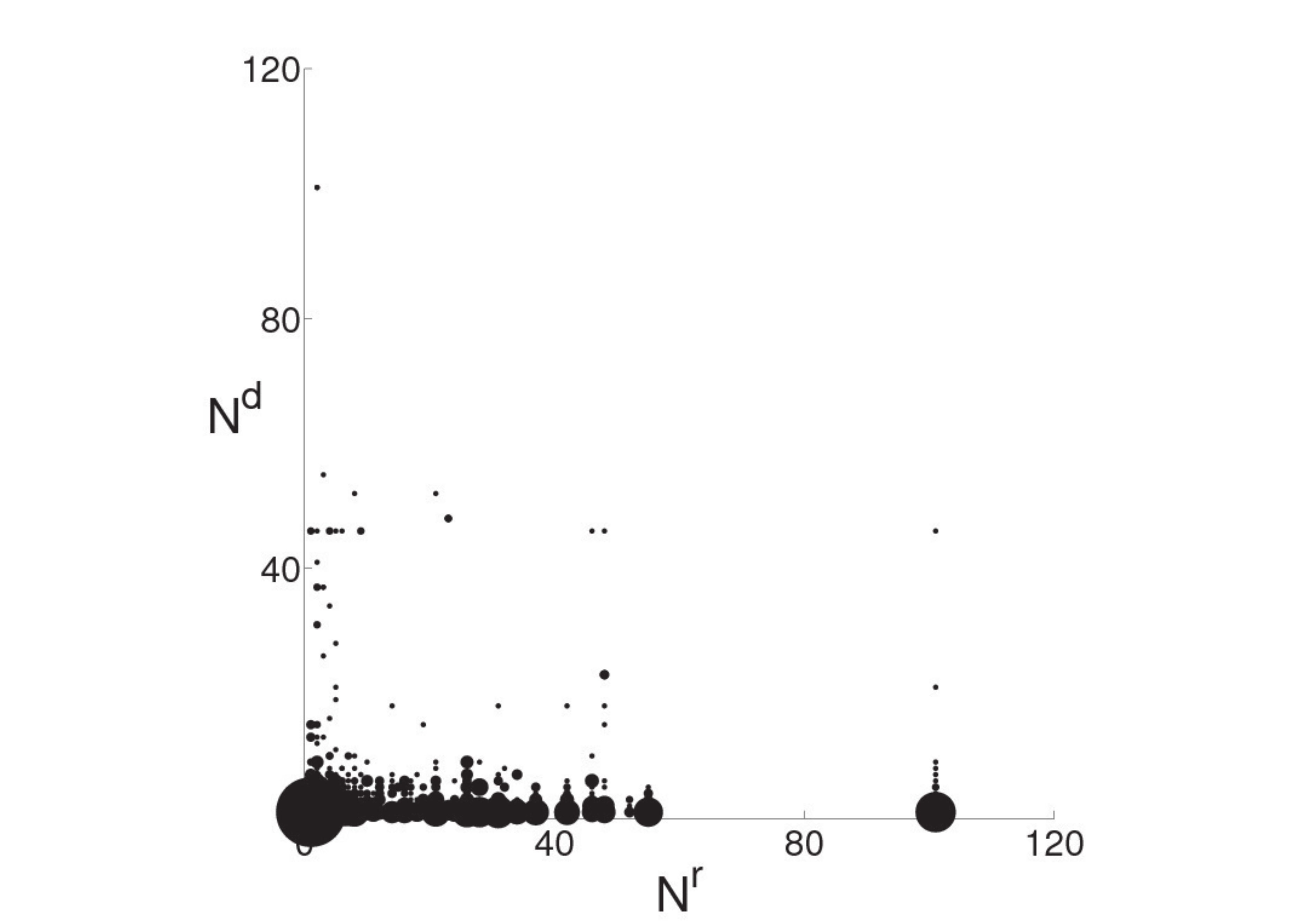}\hspace*{-6mm}\includegraphics[width=52mm]{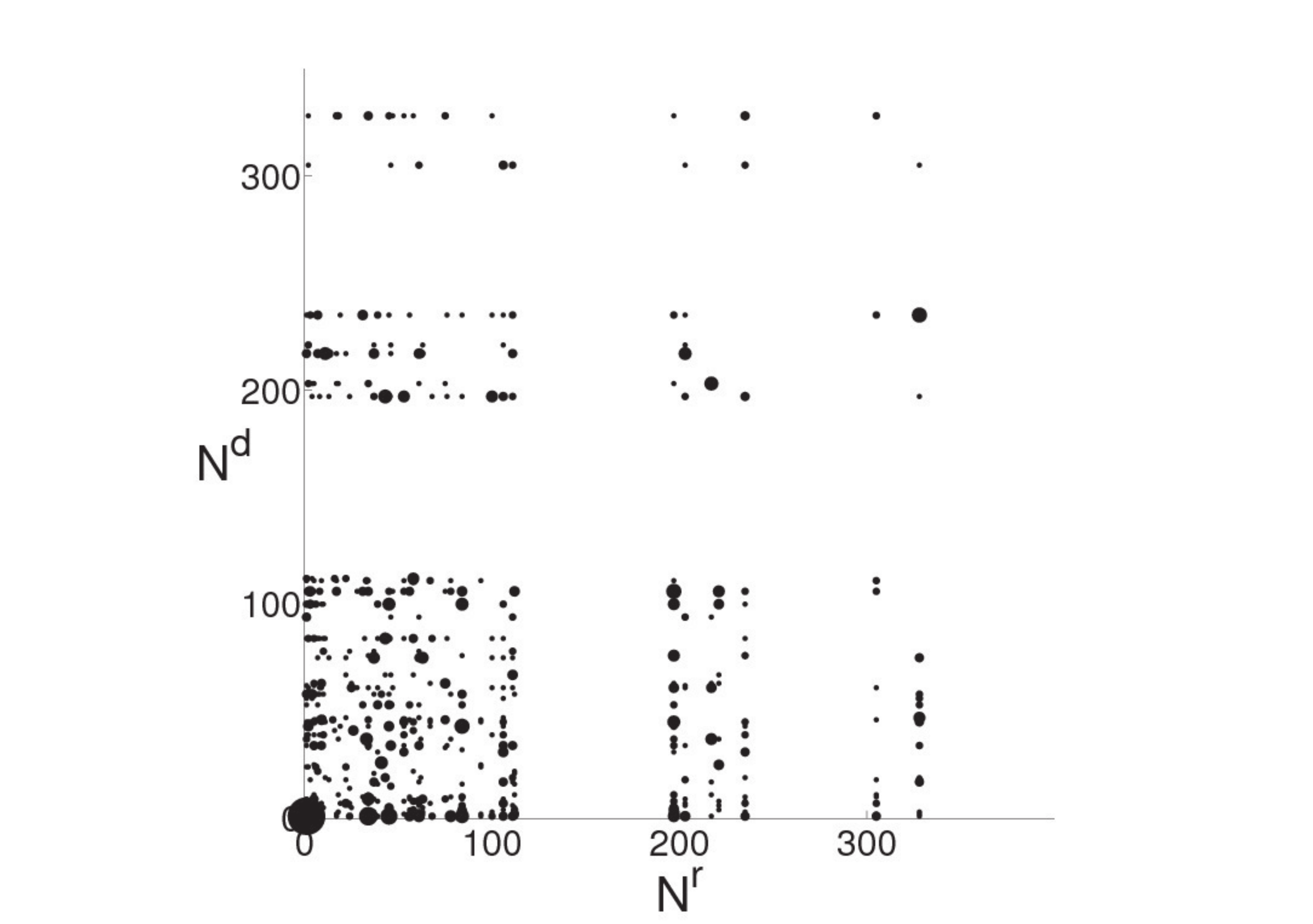}
\caption{Revert maps of the articles {\tt Benjamin Franklin} (left) and {\tt Israel
    and the apartheid analogy} (right). $N^r$ and $N^d$ are the total number of edits
  of the reverter and reverted editor respectively. The size of the mark is
  proportional to the number of reverts between them.}  \label{fig:benjamin}
\includegraphics[width=52mm]{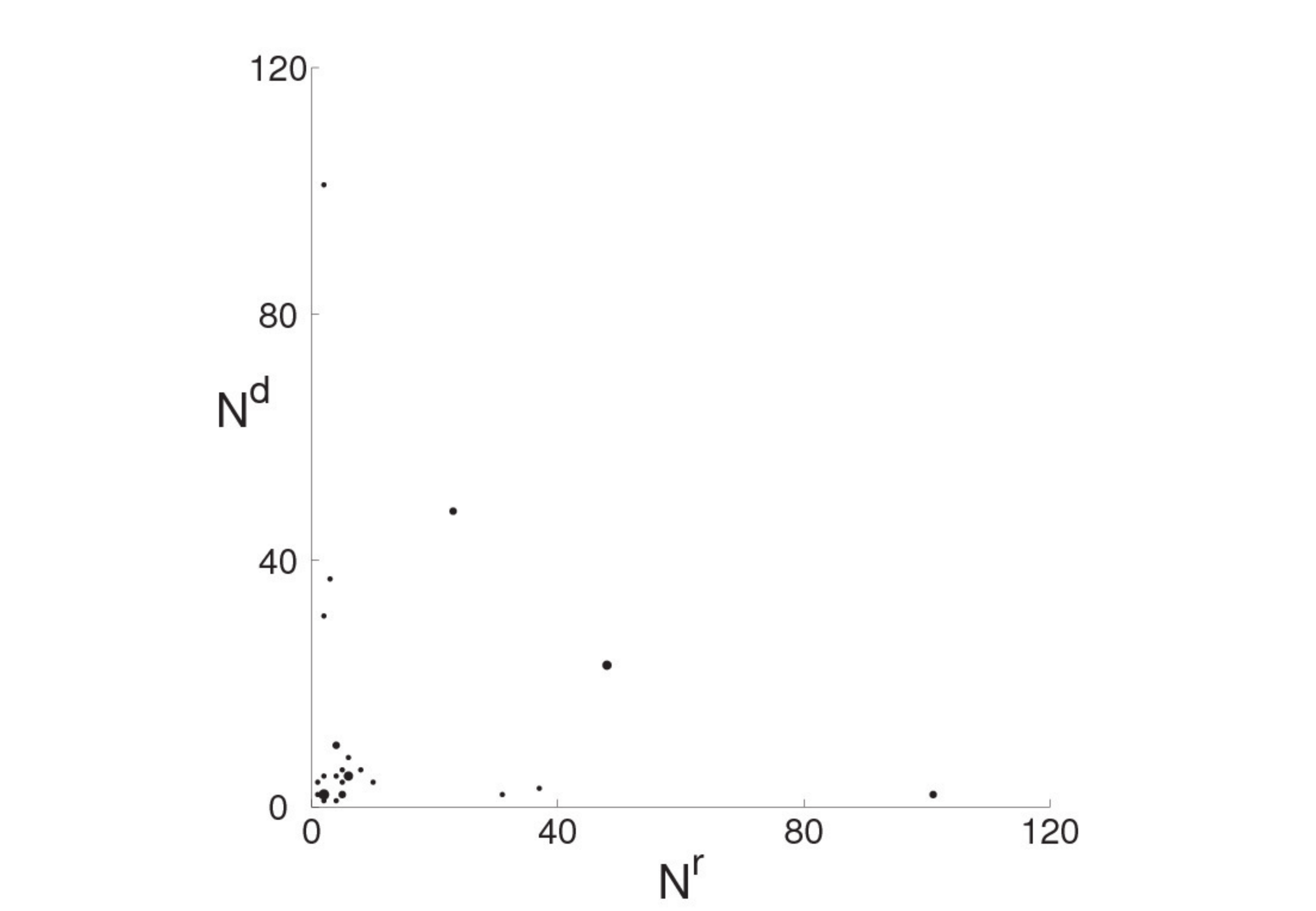}\hspace*{-6mm}\includegraphics[width=52mm]{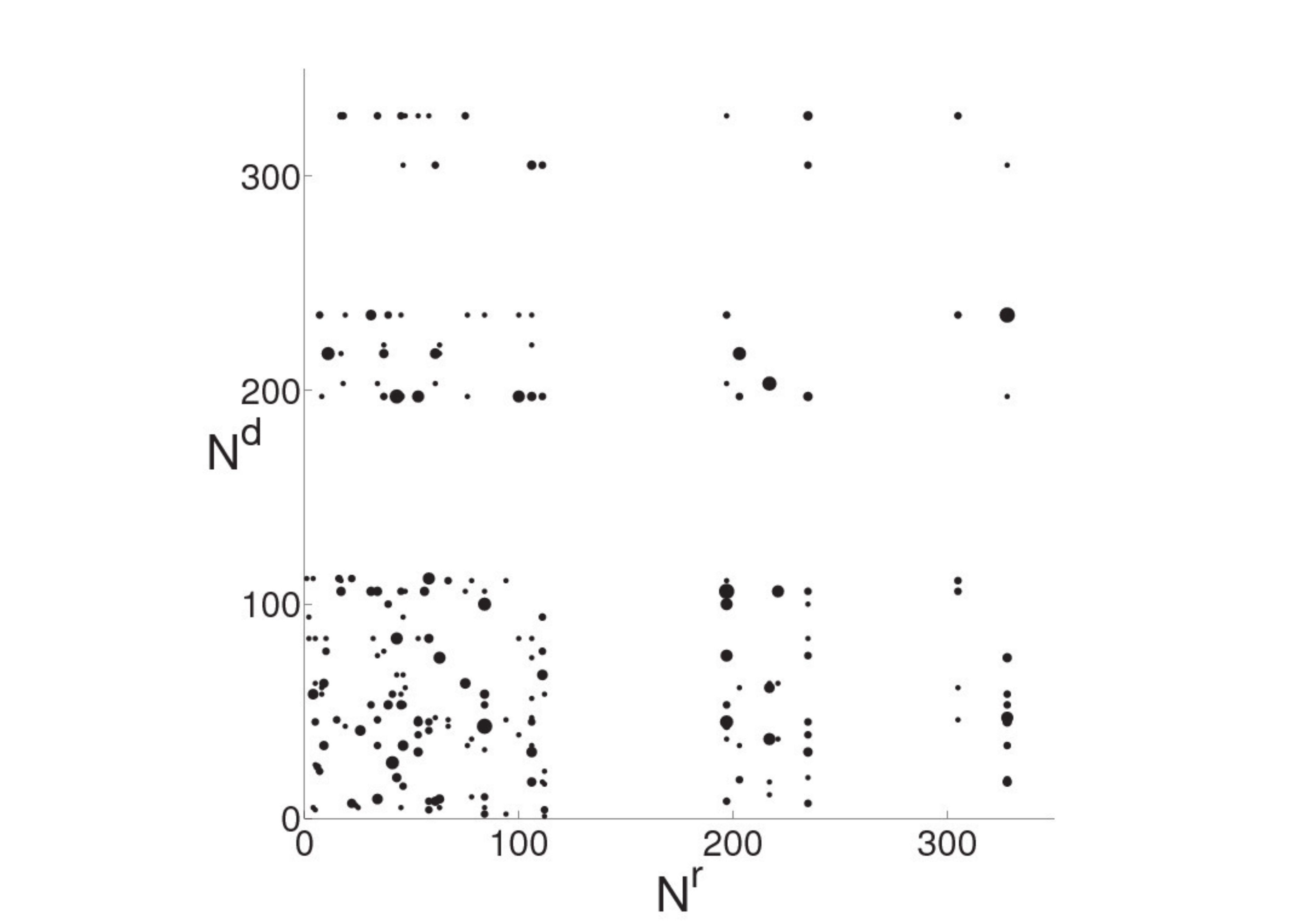}
\caption{Maps of mutual reverts in the same articles as in Fig.~\ref{fig:benjamin}.}
\label{fig:benjamin2} \end{figure}

Let us denote by $N_i$ the total number of edits in the given article of
that user who edited the revision $i$. We characterize reverts by pairs
$(N_i^d, N_j^r)$, where $r$ denotes the editor who makes the revert, and $d$
refers to the reverted editor (self-reverts are excluded).
\figurename{\ref{fig:benjamin}} represents the {\it revert map} of the
non-controversial
\R2Lurl{http://en.wikipedia.org/wiki/Benjamin_Franklin}{Benjamin Franklin} and the highly controversial
\R2Lurl{http://en.wikipedia.org/wiki/Israel_and_the_apartheid_analogy}{Israel and the apartheid analogy} articles. Each mark corresponds to one or
more reverts. The coordinates of the marks are the total number of edits of
the reverter ($N^r$) and the reverted editor ($N^d$). Clearly, the disputed
article contains more reverts between editors having large edit numbers than
the uncontroversial article.

The revert maps already distinguish disputed and
non-disputed articles, and
we can improve the results by considering only the cases, in which two editors revert each other mutually, hereafter called {\it
  mutual} reverts. This causes little change in disputed articles (compare the
right panels of \figurename{\ref{fig:benjamin}} to that of
\figurename{\ref{fig:benjamin2}}), but has great impact on non-disputed
articles (compare left panels).

Based on the rank (total edit number within an article) of editors, two main
revert types can be distinguished: when one or both of the editors have few
edits to their credit (these are typically reverts of vandalism since vandals
do not get a chance to achieve a large edit number as they get banned by
experienced users) and when both editors are experienced (created many edits).  In order to express
this distinction numerically, we use the {\it lesser} of the coordinates $N^d$
and $N^r$, so that the total count includes vandalism-related reverts as well,
but with a much smaller weight. Thus we define our raw measure of
controversiality as

\begin{displaymath}
M_r = \sum_{(N_i^d, N_j^r)} min(N_i^d, N_j^r)
\end{displaymath}

Once we developed our first autodetection algorithm based on $M_r$, we
iteratively refined the controversial and the noncontroversial seeds on
multiple languages by manually checking pages scoring very high or very low.
In this process, we improved $M_r$ in two ways: first, by multiplying with the
number of editors $E$ who ever reverted mutually  (the larger the armies, the larger the war) and define $M_i=E\times M_r$
and second, by
censuring the topmost mutually reverting editors (eliminating cases with conflicts between two persons
only). Our final measure of controversiality $M$ is thus defined by

\begin{equation}
M = E \times \sum_{(N_i^d, N_j^r) < max} min(N_i^d, N_j^r).
\end{equation}

One conceptually easy (but in practice very labor-intensive) way to validate
$M$ is by simply taking samples at different $M$ values and counting how many
pages are found.  As can be seen from Table~\ref{table:cnw}, in a binary manual
classification between $n$oncontroversial and $c$ontroversial pages
the number of controversial pages increases monotonically with
increased $M$ from a low of 27\% war to a high of 97\% as $M$ goes from 50 to
31,000. (We note here that some of the truly controversial pages such as
\R2Lurl{http://en.wikipedia.org/wiki/Anarchism}{Anarchism} have $M$
in excess of 10$^7$.)

\begin{table}[!ht]
\begin{center}
 \caption{ For a given level of $M$, number of
  $n$oncontroversial, $c$ontroversial pages in random samples of 30, $T$otal
  number of pages with greater $M$, \% of controversial pages, estimated total number $C$ of controversial pages
  with greater $M$ (numbers are in kilo and rounded to two significant digits).}
\begin{tabular}{rrrrrrrrr}
$M$ & $n$ & $c$  & $T$ & $\%c$ & $C$ (k) \\
\hline
50 & 22 & 8 & 44037 & 27  & 24 \\
100 & 16 & 14 & 34112 & 47 & 20\\
180 & 15 & 15 & 26912 & 50 & 17\\
320 & 14 & 16 & 20763 & 53 & 13 \\
560 & 14 & 16 & 15683 & 53 & 11 \\
1000 & 12 & 18 & 11732 & 60 & 9\\
5600 & 1 & 29 & 4314 & 97 & 4 \\
31000 & 1 & 29 & 1368 & 97& 1 \\
\end{tabular}
\label{table:cnw}
\end{center}
\end{table}

We have checked this measure for six different languages (eventually leaving
Romanian out for lack of data) and concluded that its overall performance is
superior to other measures, including the presence of tags marking
controversiality (see Section~\ref{sec:other}).

\section{Other indicators  of controversiality}\label{sec:other}

There are many plausible candidates for measuring controversiality, such as
the number of edits, the number of (mutual) reverts, the number of `controversial' tags, and of course variants of our own
measure $M$.  Table~\ref{table:m} compares the various methods for precision in the top 30
for each language (except Romanian for paucity of controversial pages), and
for each method considered.

\begin{table}
\begin{center}
\caption{Precision of controversiality
detection in the top 30 based on number of edits \#e, reverts \#r, mutual
reverts \#mr, raw $M_r$, $M_i$, article tag count TC, and $M$.}
\begin{tabular}{llllllll}
WP  & \#e & \#r & \#mr & $M_r$  & $M_i$ &TC & $M$ \\
\hline
{\tt cs }  & 14  & 18    & 26   & 25 & 27 & 27 & 28 \\
{\tt en }  & 27  & 29    & 29   & 26 & 28 & 30 & 28 \\
{\tt hu }  & 4   & 27    & 28   & 23 & 29 & 24 & 30 \\
{\tt fa }  & 24  & 28    & 26   & 29 & 29 & 25 & 28 \\
{\tt es }  & 23  & 26    & 29   & 27 & 28 & 28 & 29 \\
\hline
\%av     & 61 &  85 & 92 & 87 & 94 & 89 &95\\
\end{tabular}
\label{table:m}
\end{center}
\end{table}

It comes as no great surprise that the roughest measures such as the number of
edits are rather poor classifiers and `controversial' tag count (TC) are quite good in the English WP.
Unfortunately, these
measures fail to generalize from English and Spanish to smaller WPs, and given
the cultural differences, there is no assurance that as the smaller WPs mature
these measures will become increasingly applicable. As it is, measure $M$
given by Eq.~1 above reduces the precision errors of the hitherto best
classifier by over 50\% (10 errors compared to the best known method. article
tag count, which had 24). Of our own methods, the final $M$-based classifier
improves upon the initial method (counting mutual reverts, 15 errors) by a
full third.

\begin{figure}[h!]
\begin{center}
 \includegraphics[width=0.45\textwidth]{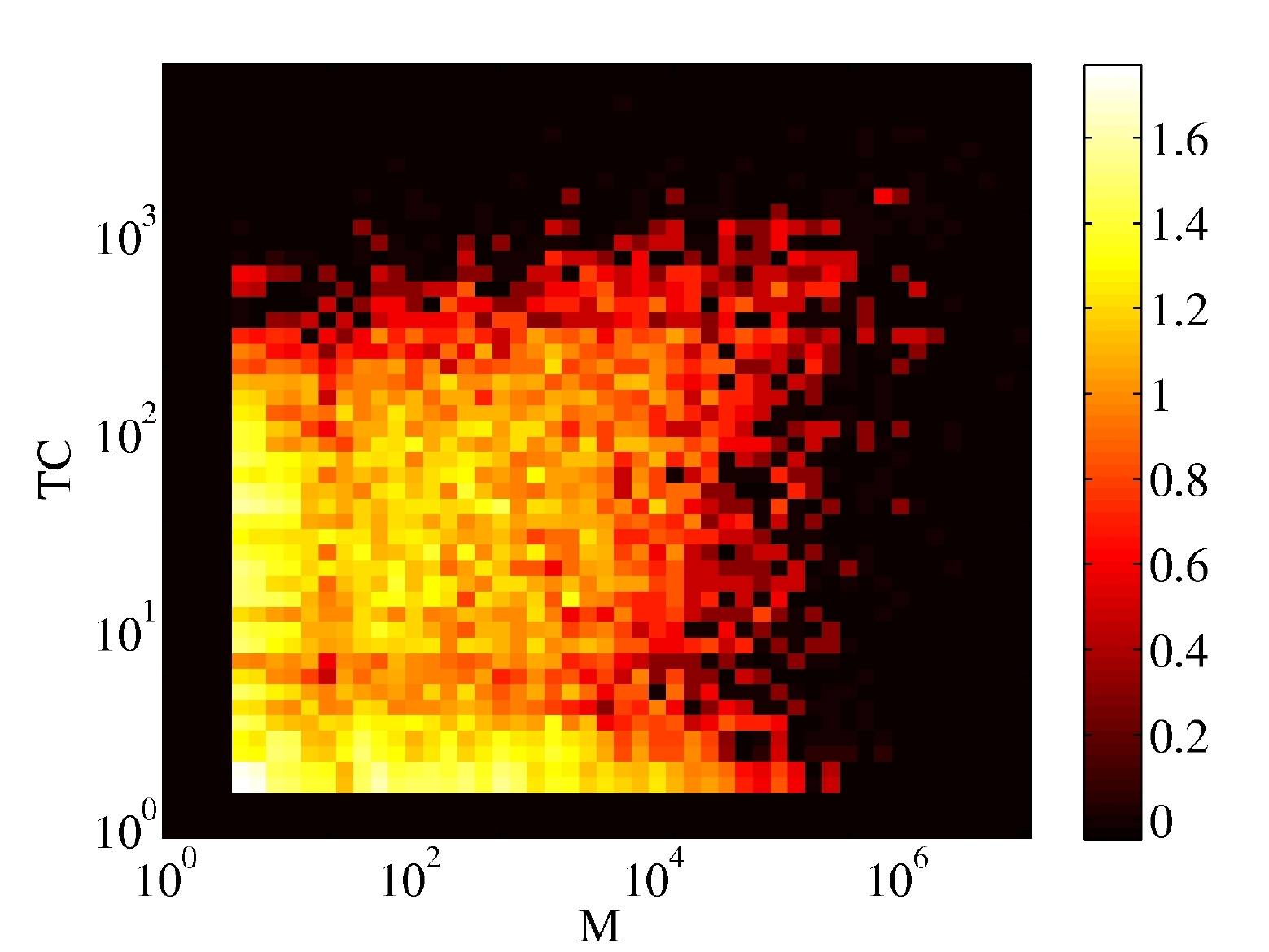}
 \caption{Scatter plot of article Tag Count (TC) vs. $M$. Color coding is proportional to the logarithm
of the number of articles lying within each cell}\label{figure:m-atc} \end{center} \end{figure}

The true value of a classifier of course depends not just on precision, but
also on recall. This is much harder to measure, since the bottom 90\% of the
sample is uncontroversial by any measure, and it would take tens of thousands
of manual judgments on random samples to obtain reliable recall figures.
Also, the threshold for large WPs is much lower than one could surmise from
inspecting the top 30 pages, for example in English $M$ is 1,620,378 for the
30th most controversial page. We selected $M > 1,000$ for cutoff, which
yields a very high controversy population, but if we were truly intent on
optimizing the threshold we would probably go down to about $M  = 200$ to see as much
vandalism as pure edit warring.  As a shortcut, we therefore plotted the
second best classifier against $M$ (see Fig.~\ref{figure:m-atc}), and sampled from the
quadrants where they make different predictions.

On the whole, articles with low tag count but high $M$ appear to be quite
controversial, even if the participants themselves fail to tag the article for
controversy. The opposite situation, with low $M$ and high TC, is found very
rarely. Besides an inherently lower precision and recall, there are some
mechanical reasons why counting controversial tags is not a perfect method to
detect disputes. There are many dispute-related tags and one has to decide
which tags to count on a per-language basis.  Page ~\cite{disptags} contains
all disputed tags, but some may indicate more serious conflict than others
(for example compare \{\{Unreferenced\}\} to \{\{Disputed\}\}).

The limitations of the earlier proposals such as \cite{Kittur2007} and
\cite{Voung2008} are evident if we check the results. For example,
\cite{Voung2008} concluded that in the {\tt Podcast} article ``a significant
amount of dispute occurred between the two pairs of users: (a) user
210.213.171.25 and user Jamadagni; and (b) user 68.121.146.76 and user
Yamamoto Ichiro". A closer look at the article reveals that user
210.213.171.25 edited the given article only once, and his edit was a
vandalism, because he multiplied several time the text of the article,
creating a revision which was 20 times larger than the previous one. Jamadagni
simply reverted, generating this way a large number of deleted words between
them. Real, recurrent disputes cannot produce this large amount of deleted
words, therefore they remain hidden. (This is not an extreme example, user
68.121.146.76 is another vandal, who edited the article only once.)

\section{Conclusion, future directions}\label{sec:conc}

We proposed a new way to measure how disputed a WP article is. We did this
because existing models have drawbacks, and only a small fraction of WP
articles were analyzed with them. We analyzed the whole WP for different
language versions, and ranked articles according to their controversy level.
Altogether, the proposed measure $M$ fares considerably better than earlier
proposals both for precision and recall, though this fact would not be evident
to the observer restricted to the top 30 articles of the English WP. For
example, in Romanian even TC fails rather spectacularly.
Based on the results obtained by our classifier,
we conclude that in most cases, the process of development of the articles is considerably peaceful and the number of
conflict cases have been overestimated in previous works, compared to our estimation of less than 1\% of articles to be a candidate for
a serious conflict.
Besides being a
robust language- and culture-independent classifier, our method also yields a
numerical ranking, which agrees well with human judgment.

While our method does well in separating out edit wars from vandalism at the
high end, much work remains to be done for lower $M$. Researchers interested
in a better controversiality measure may go back to the other indicators of
controversiality and mix these into the measure: the largest limiting factor
is the number of manually truthed examples one is willing to create for
training and testing, but if resources are pooled across teams or the
judgement task could be automated (e.g. by the Mechanical Turk) this
limitation can be overcome.

Our future goals include detection of pure (non-war-like) vandalism, a task
made all the more important by the high degree of vandalism we see.  Another
goal is the prediction of impending edit wars by monitoring the dynamics of
$M$ -- once a reasonable predictor is provided it will be possible to tag
pages for impending conflict by robots.

\section*{Acknowledgment}

This work was supported by the EU's 7th Framework
Programs FET-Open within ICTeCollective project no. 238597. Kornai also
acknowledges support from OTKA grants \#77476 (Algebra and algorithms) and
\#82333 (Semantic language technologies).  Special thanks to Santo Fortunato
for discussions and his help with data at early stages of this work.


\begin{thebibliography}{10}

\bibitem{schneider:2010}
Schneider, Jodi and Passant, Alexandre and Breslin, John
\newblock A Qualitative and Quantitative Analysis of How Wikipedia Talk Pages Are Used
\newblock {\em Proceedings of WebSci10} 26-27 April, Raleigh, NC, 2010
\newblock http://journal.webscience.org/373

\bibitem{Ratkiewicz2010}
J. Ratkiewicz, S. Fortunato, A. Flammini, F. Menczer and A. Vespignani
\newblock { Characterizing and modeling the dynamics of online popularity}
\newblock {Physical Review Letters} 105, article no. 158701, 2010.

\bibitem{Kittur2007}
Aniket Kittur, Bongwon Suh, Bryan~A. Pendleton, and Ed~H. Chi.
\newblock {He says, she says: conflict and coordination in Wikipedia}.
\newblock In {\em CHI '07: Proceedings of the SIGCHI conference on Human
  factors in computing systems}, pages 453--462, New York, NY, USA, 2007. ACM.

\bibitem{Voung2008}
Ba-Quy Vuong, Ee-Peng Lim, Aixin Sun, Minh-Tam Le, and Hady~Wirawan Lauw.
\newblock On ranking controversies in wikipedia: models and evaluation.
\newblock In {\em Proceedings of the international conference on Web search and
  web data mining}, WSDM '08, pages 171--182, New York, NY, USA, 2008. ACM.

\bibitem{Suh2007}
Bongwon Suh, E.H. Chi, B.A. Pendleton, and A.~Kittur.
\newblock Us vs. them: Understanding social dynamics in wikipedia with revert
  graph visualizations.
\newblock {\em Visual Analytics Science and Technology, 2007. VAST 2007. IEEE
  Symposium on}, pages 163--170, 30 2007-Nov. 1 2007.

\bibitem{Brandes2008}
Ulrik Brandes and J\"{u}rgen Lerner.
\newblock Visual analysis of controversy in user-generated encyclopedias.
\newblock {\em Information Visualization}, 7:34--48, March 2008.

\bibitem{Brandes2009}
Ulrik Brandes, Patrick Kenis, J\"{u}rgen Lerner, and Denise van Raaij.
\newblock Network analysis of collaboration structure in wikipedia.
\newblock In {\em Proceedings of the 18th international conference on World
  wide web}, WWW '09, pages 731--740, New York, NY, USA, 2009. ACM.

\bibitem{west2010}
Andrew~G. West, Sampath Kannan, and Insup Lee.
\newblock Detecting wikipedia vandalism via spatio-temporal analysis of
  revision metadata?
\newblock In {\em Proceedings of the Third European Workshop on System
  Security}, EUROSEC '10, pages 22--28, New York, NY, USA, 2010. ACM.

\bibitem{adler2010} B. Thomas Adler, Luca de Alfaro, and Ian Pye.
\newblock Detecting Wikipedia Vandalism using WikiTrust: Lab Report
for PAN at CLEF 2010 \newblock In {\em Notebook Papers of CLEF 2010
LABs and Workshops,} 22-23 September, Padoa, Italy, 2010. ISBN
978-88-904810-0-0.

\bibitem{sumi2011}
R. Sumi, T. Yasseri, A. Rung, A. Kornai, and J. Kert\'esz.
\newblock Characterization and prediction of Wikipedia edit wars
\newblock In {\em Proceedings of the ACM WebSci'11} June 14-17, Koblenz, Germany, 2011.

\bibitem {disptags}
{\tt http://en.wikipedia.org/wiki/Wikipedia:Template$\_$} {\tt
  messages/Disputes}



\end{thebibliography}
\end{document}